%% file: paper_maep.tex
\documentclass{article}

\def\nonanonymous{}

\ifdefined\nonanonymous
\usepackage[final]{neurips_2019}
\else
\usepackage{neurips_2019}
\fi

\usepackage[utf8]{inputenc} 
\usepackage[T1]{fontenc}    
\usepackage{hyperref}       
\usepackage{url}            
\usepackage{booktabs}       
\usepackage{amsfonts}       
\usepackage{nicefrac}       
\usepackage{microtype}      
\usepackage{wrapfig}

\usepackage{amssymb}
\usepackage{amsthm}
\usepackage{bm}
\usepackage{mathtools}
\usepackage{caption}
\usepackage{subcaption}
\usepackage{caption}
\usepackage{csquotes}
\usepackage{layouts}
\usepackage{float}
\usepackage{todonotes}
\usepackage{enumitem}
\usepackage{lipsum}

\usepackage{algorithm}
\usepackage[noend]{algpseudocode}
\algnewcommand{\Let}[2]{\State #1 $\gets$ #2}
\algrenewcommand\Call[2]{\textproc{#1}(#2)}

\usepackage{multirow}
\usepackage{tabu}
\let\cite\citep
\title{Measuring Arithmetic Extrapolation Performance}

\author{%
  Andreas Madsen \\
  Computationally Demanding \\
  \texttt{amwebdk@gmail.com}
  \And
  Alexander Rosenberg Johansen \\
  Technical University of Denmark \\
  \texttt{aler@dtu.dk} \\
}

\begin{document}

\maketitle


\begin{abstract}
The Neural Arithmetic Logic Unit (NALU) is a neural network layer that can learn exact arithmetic operations between the elements of a hidden state.
The goal of NALU is to learn perfect extrapolation, which requires learning the exact underlying logic of an unknown arithmetic problem.
Evaluating the performance of the NALU is non-trivial as one arithmetic problem might have many solutions.
As a consequence, single-instance MSE has been used to evaluate and compare performance between models.
However, it can be hard to interpret what magnitude of MSE represents a correct solution and models sensitivity to initialization.
We propose using a success-criterion to measure if and when a model converges.
Using a success-criterion we can summarize success-rate over many initialization seeds and calculate confidence intervals.
We contribute a generalized version of the previous arithmetic benchmark to measure models sensitivity under different conditions.
This is, to our knowledge, the first extensive evaluation with respect to convergence of the NALU and its sub-units.
Using a success-criterion to summarize 4800 experiments we find that consistently learning arithmetic extrapolation is challenging, in particular for multiplication.
\ifdefined\nonanonymous\footnote{code for experiments is publicly available at: \url{https://github.com/AndreasMadsen/stable-nalu}.}\fi
\end{abstract}
\input{sections/introduction}
\input{sections/related-work}
\input{sections/methods}
\input{sections/results}

\input{sections/conclusion}
\clearpage
\ifdefined\nonanonymous
\subsubsection*{Acknowledgments}
We would like to thank Andrew Trask and the other authors of the NALU paper, for highlighting the importance and challenges of extrapolation in Neural Networks. We would also like to thank the students Raja Shan Zaker Kreen and William Frisch Møller from The Technical University of Denmark, who initially showed us that the NALU does not converge consistently.

This research is funded by the Innovation Foundation Denmark through the DABAI project.
\fi

\bibliographystyle{plainnat}
\bibliography{bibliography}

\newpage
\appendix
\input{appendix/experimental-details}

\end{document}

%% file: sections/introduction.tex
\section{Introduction}

When using neural networks to learn simple arithmetic problems, such as counting, multiplication, or comparison they systematically fail to extrapolate onto unseen ranges \cite{stillNotSystematic,suzgun2019evaluating,trask-nalu}.
The absence of inductive bias makes it difficult for neural networks to extrapolate well on arithmetic tasks as they lack the underlying logic to represent the required operations.

A recently proposed model, called NALU \cite{trask-nalu}, attempts to solve the problem of arithmetic extrapolation.
However, for arithmetic extrapolation there are no broadly accepted guidelines for evaluating model performance.
As a result, single-instance MSE is used for comparison.

As exact extrapolation requires correctly solving a logical problem we advocate that the performance metrics of interest should be: 1) has it learned the underlying logic, 2) how often does it learn the correct solution, and 3) how fast does it converge?

Motivated by these questions we propose using a success-criterion to determine if the underlying logic has been learned.
We measure success-rate and provide a binomial confidence interval by initializing and training the NALU over multiple seeds.
For each seed, we use the first iteration that satisfy the success-criterion to measure when the model has succeeded.
As the success-criterion is based on an MSE divergence from an optimal solution it can be generalized to any model.

Finally, we propose and report a sparsity measurement for models that satisfy the success-criterion.
Sparsity of the parameters has previously been emphasized as important for a correct solution \cite{trask-nalu}.

%% file: sections/related-work.tex
\section{Related work}
\citet{GridLSTM,lte,NeuralGPU,FreivaldsL17} solves integer arithmetic operations as a classification task and reports exact match accuracy.
Using accuracy is useful for well-defined classification tasks, but is hard to use for real number regression problems.
Our criterion mimics exact match by defining an MSE $\epsilon$-threshold.

%% file: sections/methods.tex
\section{Simple Function Learning Tasks}
\begin{figure}[t]
\centering
\includegraphics[width=\linewidth,trim={0 0.2cm 0 0.2cm},clip]{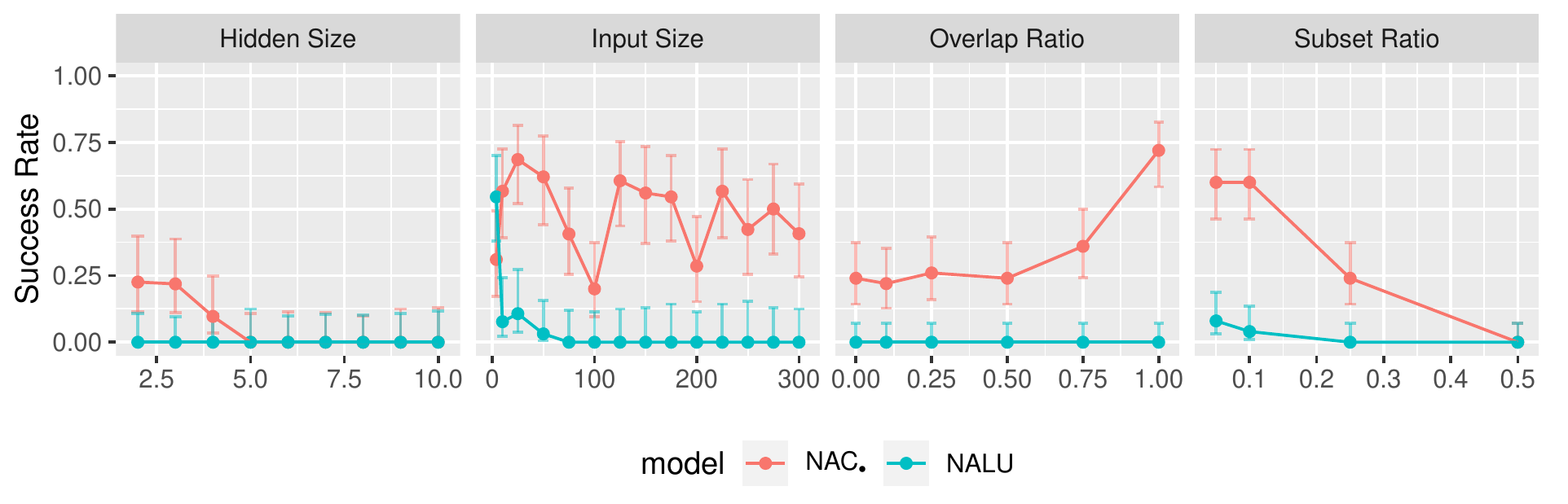}
\caption{Shows success-rate given different dataset parameters and the models hidden-size using the multiplication operation. Means are over 50 different seeds, with 95\% confidence intervals.} 
\label{fig:simple-function-task-parameters}
\end{figure}

The ``Simple Function Learning Tasks'' is a synthetic dataset that tests arithmetic extrapolation.
The problem is defined as summing two random subsets of $\mathbf{x}\in \mathbb{R}^d$ followed by an arithmetic operation $\circ \in \{+, -, \times, \div\}$ on these sums.
Extrapolation can be tested by modifying the sampling range of $\mathbf{x}$.

\begin{algorithm}[h]
  \caption{Dataset sampling algorithm. Default values are specified for input-size ($d$), subset-ratio ($s$), and overlap-ratio ($o$). Default interpolation range is $[1,2]$ and default extrapolation range is $[2,6]$.}
  \begin{algorithmic}[1]
    \Function{Dataset}{${\Call{Op}{\cdot, \cdot}: \mathrm{Operation}}$, ${R: \mathrm{Range}}$, ${d=100}$, ${s=0.25}$, ${o=0.5}$}
      \Let{$\mathbf{x}$}{\Call{Uniform}{$R_{lower}, R_{upper}, i$}} \Comment{Sample $d$ elements uniformly}
      \Let{$k$}{\Call{Uniform}{$0, 1 - 2s - o$}} \Comment{Sample offset. Same for interpolation and extrapolation.}
      \Let{$a$}{\Call{Sum}{$\mathbf{x}[dk:d(k+s)]$}} \Comment{Create sum $a$ from a subset of length $s \cdot d$}
      \Let{$b$}{\Call{Sum}{$\mathbf{x}[d(k+s-o):d (k+2s-0)]$}} \Comment{Create sum $b$ from a subset of length $s \cdot d$}
      \Let{$t$}{\Call{Op}{$a, b$}} \Comment{Perform operation on $a$ and $b$}
      \State \Return{$\mathbf{x}, t$}
    \EndFunction
  \end{algorithmic}
  \label{alg:simple-function-task-dataset-sampling}
\end{algorithm}

Solving the task on extrapolation requires learning the underlying logic of arithmetic operations from the training range.
As logic is discrete, a solution to the problem is either correct or wrong.

To evaluate a solution we propose comparing the MSE, of the entire testset, to the MSE of a nearly-perfect solution on the extrapolation range.
The nearly-perfect solution is defined as performing the operation perfectly, but allowing a small error in the sum-of-subsets (line 4 and 5 in {Algorithm~\ref{alg:simple-function-task-dataset-sampling}}).
This threshold can be simulated with $\frac{1}{N} \sum_{i=1}^{N} (\mathrm{Op}(\mathbf{W}^\epsilon_{1,:} \mathbf{x}_i, \mathbf{W}^\epsilon_{2,:} \mathbf{x}_i) - t_i)^2$ for $N = 1000000$, where $\mathbf{W}^\epsilon = \mathbf{W}^* \pm \epsilon$ and $\mathbf{W}^*$ is the perfect $\mathbf{W}$ required to compute the optimal solution. We set $\epsilon = 10^{-5}$.

Using a success-criterion has the advantage of being more interpretable, models that failed to converge will not obscure the mean, and as the number of successes will follow a binomial distribution we can calculate a confidence interval \cite{wilson-binomial}.

With a success-criterion we can evaluate when a model succeeds.
Since this metric cannot be negative, we model the confidence interval with a gamma distribution and report a 95\% confidence intervals of the mean, by using maximum likelihood profiling.

Finally, the parameters of the NALU are argued to be ``biased to be close to -1, 0, -1'' \cite{trask-nalu}.
We propose to measure a sparsity error of the NALU parameters with $\max_i \min(|W_i|, |1 - |W_i||)$.
As the sparsity error is between $[0, 0.5]$ we use a modified beta distribution with support in $[0, 0.5]$ and report a 95\% confidence interval of the mean, by using maximum likelihood profiling.

The choice of gamma and beta distribution may not be perfect.
However, a normal distribution would be problematic when the mean is close to the bounds, as it will have a large probability mass outside of the support bounds and thus provide inaccurate confidence intervals.

%% file: sections/results.tex
\section{Results}

\begin{figure}[h]
\centering
\includegraphics[width=\linewidth,trim={0 1.6cm 0 0.2cm},clip]{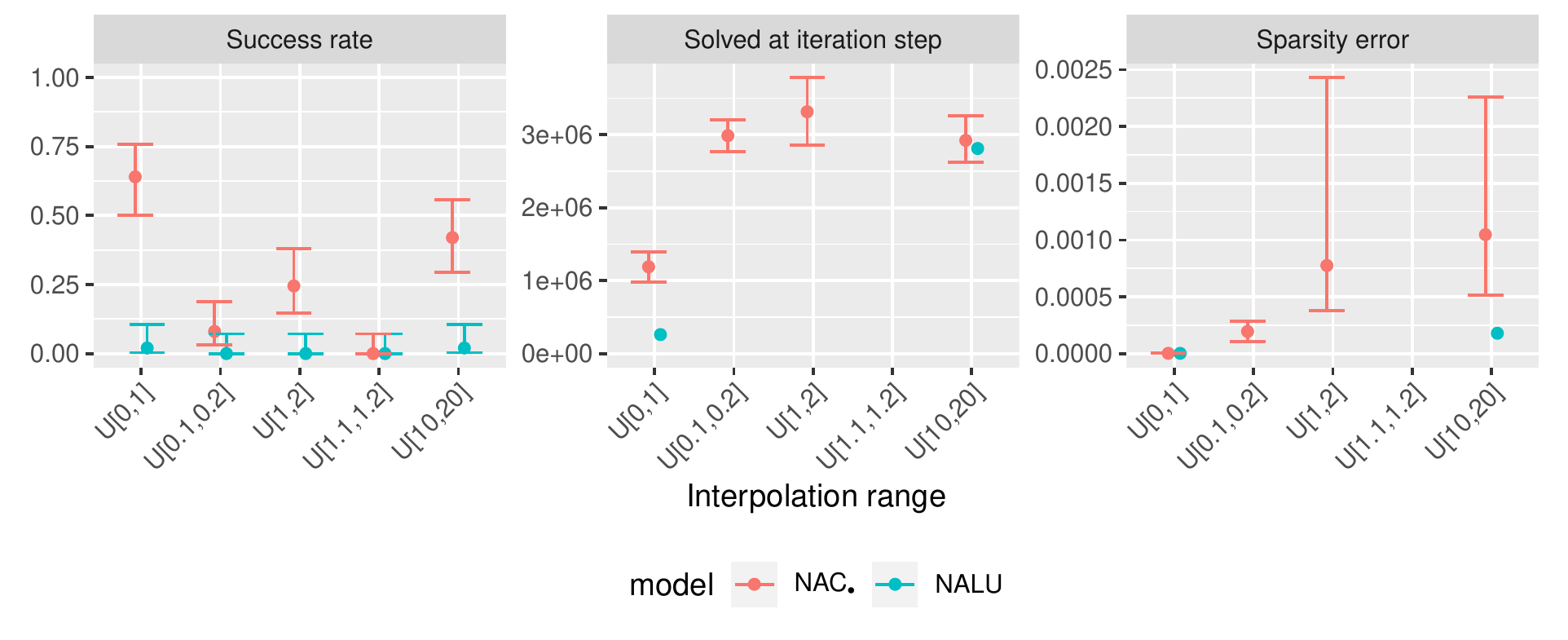}
\caption{Shows success-rate, when models converged, and sparsity error for the multiplication operation. Means are over 50 seeds. We provide experimental details in Appendix \ref{appendix:experimential-detials}.}
\label{fig:simple-function-task-range}
\end{figure}

\input{results/function_task_static_reproduce.tex}

%% file: results/function_task_static_reproduce.tex
\begin{table}[h]

\caption{\label{tab:function-task-static-defaults}Shows success-rate, when models converged, and sparsity error. Means are over 100 seeds.}
\centering
\begin{tabular}{crlll}
\toprule
\multicolumn{1}{c}{Op} & \multicolumn{1}{c}{Model} & \multicolumn{1}{c}{Success Rate} & \multicolumn{1}{c}{Solved at} & \multicolumn{1}{c}{Sparsity error} \\
\midrule
 & $\mathrm{NAC}_{\bullet}$ & $\mathbf{31\%} {~}^{+10\%}_{-8\%}$ & $\mathbf{3.0 \cdot 10^{6}} {~}^{+2.9 \cdot 10^{5}}_{-2.4 \cdot 10^{5}}$ & $\mathbf{5.8 \cdot 10^{-4}} {~}^{+4.8 \cdot 10^{-4}}_{-2.6 \cdot 10^{-4}}$\\

\nopagebreak
\multirow{-2}{*}{\centering\arraybackslash $\bm{\times}$} & NALU & $0\% {~}^{+4\%}_{-0\%}$ & --- & ---\\
\cmidrule{1-5}
 & $\mathrm{NAC}_{\bullet}$ & $\mathbf{0\%} {~}^{+4\%}_{-0\%}$ & --- & ---\\

\nopagebreak
\multirow{-2}{*}{\centering\arraybackslash $\bm{\mathbin{/}}$} & NALU & $\mathbf{0\%} {~}^{+4\%}_{-0\%}$ & --- & ---\\
\cmidrule{1-5}
 & $\mathrm{NAC}_{+}$ & $\mathbf{100\%} {~}^{+0\%}_{-4\%}$ & $4.9 \cdot 10^{5} {~}^{+5.2 \cdot 10^{4}}_{-4.5 \cdot 10^{4}}$ & $2.3 \cdot 10^{-1} {~}^{+6.5 \cdot 10^{-3}}_{-6.5 \cdot 10^{-3}}$\\

\nopagebreak
 & Linear & $\mathbf{100\%} {~}^{+0\%}_{-4\%}$ & $\mathbf{6.3 \cdot 10^{4}} {~}^{+2.5 \cdot 10^{3}}_{-3.3 \cdot 10^{3}}$ & $2.5 \cdot 10^{-1} {~}^{+3.6 \cdot 10^{-4}}_{-3.6 \cdot 10^{-4}}$\\

\nopagebreak
\multirow{-3}{*}{\centering\arraybackslash $\bm{+}$} & NALU & $14\% {~}^{+8\%}_{-5\%}$ & $1.6 \cdot 10^{6} {~}^{+3.8 \cdot 10^{5}}_{-3.3 \cdot 10^{5}}$ & $\mathbf{1.7 \cdot 10^{-1}} {~}^{+2.7 \cdot 10^{-2}}_{-2.5 \cdot 10^{-2}}$\\
\cmidrule{1-5}
 & $\mathrm{NAC}_{+}$ & $\mathbf{100\%} {~}^{+0\%}_{-4\%}$ & $\mathbf{3.7 \cdot 10^{5}} {~}^{+3.8 \cdot 10^{4}}_{-3.8 \cdot 10^{4}}$ & $2.3 \cdot 10^{-1} {~}^{+5.4 \cdot 10^{-3}}_{-5.4 \cdot 10^{-3}}$\\

\nopagebreak
 & Linear & $7\% {~}^{+7\%}_{-4\%}$ & $1.4 \cdot 10^{6} {~}^{+7.0 \cdot 10^{5}}_{-6.1 \cdot 10^{5}}$ & $\mathbf{1.8 \cdot 10^{-1}} {~}^{+7.2 \cdot 10^{-2}}_{-5.8 \cdot 10^{-2}}$\\

\nopagebreak
\multirow{-3}{*}{\centering\arraybackslash $\bm{-}$} & NALU & $14\% {~}^{+8\%}_{-5\%}$ & $1.9 \cdot 10^{6} {~}^{+4.4 \cdot 10^{5}}_{-4.5 \cdot 10^{5}}$ & $2.1 \cdot 10^{-1} {~}^{+2.2 \cdot 10^{-2}}_{-2.2 \cdot 10^{-2}}$\\
\bottomrule
\end{tabular}
\end{table}

%% file: sections/conclusion.tex
\section{Conclusion}
We provide the most extensive study of the Neural Arithmetic Logic Unit to date using a generalized version of the ``Simple Function Learning Tasks''.
Our study, through varying task complexities, evaluates the NALUs ability to learn the logic of arithmetic operations.

To evaluate performance on solving arithmetic operations we define a new success-criterion that approximates an exact match.
With a success-criterion we measure how often a model successfully solve the problem given different initailization seeds, a binomial confidence interval, and at what iteration the model satisfy the criterion.
Our results find that the NALU and its sub-units can require many trials to learn.
In particularly for multiplication and division.
Furthermore, we find that for subtraction and addition the solution is not always sparse.

Our results are not different from the original results, but highlights the importance of also discussing a models sensitivity to initialization.
We hope that future research will consider using success-rates as a comparison for the performance of arithmetic units.

%% file: appendix/experimental-details.tex
\section{Experimental details}
\label{appendix:experimential-detials}

\subsection{NALU definition}
The Neural Arithmetic Logic Unit (NALU) consists of two sub-units; the $\text{NAC}_{+}$ and $\text{NAC}_{\bullet}$. The sub-units represent either the $\{+, -\}$ or the $\{\times, \div \}$ operations. The NALU then assumes that either $\text{NAC}_{+}$ or $\text{NAC}_{\bullet}$ will be selected exclusively, using a sigmoid gating-mechanism.

The $\text{NAC}_{+}$ and $\text{NAC}_{\bullet}$ are defined accordingly,
\begin{align}
W_{h_\ell, h_{\ell-1}} &= \tanh(\hat{W}_{h_\ell, h_{\ell-1}}) \sigma(\hat{M}_{h_\ell, h_{\ell-1}}) \label{eq:weight}\\
\textrm{NAC}_+:\ z_{h_\ell} &= \sum_{h_{\ell-1}=1}^{H_{\ell-1}} W_{h_{\ell}, h_{\ell-1}} z_{h_{\ell-1}} \label{eq:naca}\\
\textrm{NAC}_\bullet:\ z_{h_\ell} &= \exp\left(\sum_{h_{\ell-1}=1}^{H_{\ell-1}} W_{h_{\ell}, h_{\ell-1}} \label{eq:nacm}\log(|z_{h_{\ell-1}}| + \epsilon) \right)
\end{align}
where $\hat{\mathbf{W}}, \hat{\mathbf{M}} \in \mathbb{R}^{H_{\ell} \times H_{\ell-1}}$ are weight matrices and $z_{h_{\ell-1}}$ is the input. The matrices are combined using a tanh-sigmoid transformation to bias the parameters towards a $\{-1,0,1\}$ solution. Having $\{-1,0,1\}$ allows $\text{NAC}_{+}$ to perform exact $\{+, -\}$ operations between elements of a vector.
The $\text{NAC}_{\bullet}$ uses an exponential-log transformation to create the $\{\times, \div \}$ operations (within $\epsilon$ precision).

The NALU combines these units with a gating mechanism $\mathbf{z} = \mathbf{g} \odot \text{NAC}_{+} + (1 - \mathbf{g}) \odot \text{NAC}_{\bullet}$ given $\mathbf{g} = \sigma(\mathbf{G} \mathbf{x})$. Thus allowing NALU to decide between all of the $\{+, -, \times, \div\}$ operations using backpropagation.

\subsection{Model definitions and setup}

Models are defined in table \ref{tab:simple-function-task-model-defintions} and are all optimized with Adam optimization \cite{adam-optimization} using default parameters, and trained over $5 \cdot 10^6$ iterations. Training takes about 6 hours on a single CPU core (\text{8-Core Intel Xeon E5-2665 2.4GHz}). We run 4800 experiments on a HPC cluster.

The training dataset is continuously sampled from the interpolation range where a different seed is used for each experiment, all experiments use a mini-batch size of 128 observations, a fixed validation dataset with $1 \cdot 10^4$ observations sampled from the interpolation range, and a fixed test dataset with $1 \cdot 10^4$ observations sampled from the extrapolation range.

We evaluate each metric every $1000$ iterations on the test set that uses the extrapolation range, and choose the best iteration based on the validation dataset that uses the interpolation range.

For figure \ref{fig:simple-function-task-range}, the following extrapolation ranges were used: ${\mathrm{U}[-2,-1] \rightarrow \mathrm{U}[-6,-2]}$, ${\mathrm{U}[-2,2] \rightarrow \mathrm{U}[-6,-2] \cup \mathrm{U}[2,6]}$, ${\mathrm{U}[0,1] \rightarrow \mathrm{U}[1,5]}$, ${\mathrm{U}[0.1,0.2] \rightarrow \mathrm{U}[0.2,2]}$, ${\mathrm{U}[1,2] \rightarrow \mathrm{U}[2,6]}$, ${\mathrm{U}[10, 20] \rightarrow \mathrm{U}[20, 40]}$.

\begin{table}[h]
\caption{Model definitions}
\label{tab:simple-function-task-model-defintions}
\centering
\begin{tabular}{r l l l l l}
\toprule
 Model & Layer 1 & Layer 2 \\
 \midrule
  $\mathrm{NAC}_{\bullet}$ & $\mathrm{NAC}_{+}$ & $\mathrm{NAC}_{\bullet}$ \\
 $\mathrm{NAC}_{+}$ & $\mathrm{NAC}_{+}$ & $\mathrm{NAC}_{+}$ \\
 NALU & NALU & NALU \\
 Linear & Linear & Linear \\
\bottomrule
\end{tabular}
\end{table}

%% file: paper_maep.bbl
\begin{thebibliography}{9}
\providecommand{\natexlab}[1]{#1}
\providecommand{\url}[1]{\texttt{#1}}
\expandafter\ifx\csname urlstyle\endcsname\relax
  \providecommand{\doi}[1]{doi: #1}\else
  \providecommand{\doi}{doi: \begingroup \urlstyle{rm}\Url}\fi

\bibitem[Freivalds and Liepins(2017)]{FreivaldsL17}
Karlis Freivalds and Renars Liepins.
\newblock Improving the neural {GPU} architecture for algorithm learning.
\newblock \emph{CoRR}, abs/1702.08727, 2017.
\newblock URL \url{http://arxiv.org/abs/1702.08727}.

\bibitem[Kaiser and Sutskever(2016)]{NeuralGPU}
Lukasz Kaiser and Ilya Sutskever.
\newblock Neural gpus learn algorithms.
\newblock In \emph{4th International Conference on Learning Representations,
  {ICLR} 2016, San Juan, Puerto Rico, May 2-4, 2016, Conference Track
  Proceedings}, 2016.
\newblock URL \url{http://arxiv.org/abs/1511.08228}.

\bibitem[Kalchbrenner et~al.(2016)Kalchbrenner, Danihelka, and
  Graves]{GridLSTM}
Nal Kalchbrenner, Ivo Danihelka, and Alex Graves.
\newblock Grid long short-term memory.
\newblock In \emph{4th International Conference on Learning Representations,
  {ICLR} 2016, San Juan, Puerto Rico, May 2-4, 2016, Conference Track
  Proceedings}, 2016.
\newblock URL \url{http://arxiv.org/abs/1507.01526}.

\bibitem[{Kingma} and {Ba}(2014)]{adam-optimization}
Diederik~P. {Kingma} and Jimmy {Ba}.
\newblock {Adam: A Method for Stochastic Optimization}.
\newblock In \emph{The 3rd International Conference for Learning
  Representations, San Diego, 2015}, page arXiv:1412.6980, Dec 2014.

\bibitem[Lake and Baroni(2018)]{stillNotSystematic}
Brenden~M. Lake and Marco Baroni.
\newblock Generalization without systematicity: On the compositional skills of
  sequence-to-sequence recurrent networks.
\newblock In \emph{Proceedings of the 35th International Conference on Machine
  Learning, {ICML} 2018, Stockholmsm{\"{a}}ssan, Stockholm, Sweden, July 10-15,
  2018}, pages 2879--2888, 2018.
\newblock URL \url{http://proceedings.mlr.press/v80/lake18a.html}.

\bibitem[Suzgun et~al.(2019)Suzgun, Belinkov, and
  Shieber]{suzgun2019evaluating}
Mirac Suzgun, Yonatan Belinkov, and Stuart~M. Shieber.
\newblock On evaluating the generalization of lstm models in formal languages.
\newblock In \emph{Proceedings of the Society for Computation in Linguistics
  (SCiL)}, pages 277--286, January 2019.

\bibitem[Trask et~al.(2018)Trask, Hill, Reed, Rae, Dyer, and
  Blunsom]{trask-nalu}
Andrew Trask, Felix Hill, Scott~E Reed, Jack Rae, Chris Dyer, and Phil Blunsom.
\newblock Neural arithmetic logic units.
\newblock In S.~Bengio, H.~Wallach, H.~Larochelle, K.~Grauman, N.~Cesa-Bianchi,
  and R.~Garnett, editors, \emph{Advances in Neural Information Processing
  Systems 31}, pages 8035--8044. Curran Associates, Inc., 2018.
\newblock URL
  \url{http://papers.nips.cc/paper/8027-neural-arithmetic-logic-units.pdf}.

\bibitem[Wilson(1927)]{wilson-binomial}
Edwin~B. Wilson.
\newblock Probable inference, the law of succession, and statistical inference.
\newblock \emph{Journal of the American Statistical Association}, 22\penalty0
  (158):\penalty0 209--212, 1927.
\newblock \doi{10.1080/01621459.1927.10502953}.
\newblock URL
  \url{https://www.tandfonline.com/doi/abs/10.1080/01621459.1927.10502953}.

\bibitem[Zaremba and Sutskever(2014)]{lte}
Wojciech Zaremba and Ilya Sutskever.
\newblock Learning to execute.
\newblock \emph{CoRR}, abs/1410.4615, 2014.
\newblock URL \url{http://arxiv.org/abs/1410.4615}.

\end{thebibliography}
